%% file: main.tex
\documentclass{article}

\usepackage[preprint]{neurips_2026}

\usepackage[utf8]{inputenc}
\usepackage[T1]{fontenc}
\usepackage{hyperref}
\usepackage{url}
\usepackage{booktabs}
\usepackage{amsfonts}
\usepackage{nicefrac}
\usepackage{microtype}
\usepackage{xcolor}

\usepackage{graphicx}
\usepackage{amsmath}
\usepackage{multirow}
\usepackage{wrapfig}
\usepackage{enumitem}
\usepackage{placeins}

\newcommand{\eg}{\emph{e.g.}}

\newcommand{\geminiFlash}{\textsc{Gmini3-Flash}}
\newcommand{\gpt}{\textsc{GPT-5}}

\newcommand{\qwenMiddle}{\textsc{Qwen3-VL-32B}}
\newcommand{\qwenSmall}{\textsc{Qwen3-VL-8B}}
\newcommand{\gemodels}{\textsc{GEN\_AUG}}
\newcommand{\nano}{\textsc{NanoBananaPro}}
\newcommand{\gptImage}{\textsc{GPTImage1}}
\newcommand{\qwenImage}{\textsc{QwenImageEdit}}
\newcommand{\joyomni}{\textsc{CamCraft}}

\title{Thinking with Novel Views: A Systematic Analysis of Generative-Augmented Spatial Intelligence}

\author{
  \textbf{Yanbing Zhang, Bo Wang, Jianhui Liu, Nan Jiang, Jiaxiu Jiang, Haoze Sun}\\
  \textbf{Yijun Yang, Shenghe Zheng, Lin Song, Haoyang Huang, Nan Duan, Wenbo Li\textsuperscript{\textdagger}}\\
  \normalfont Joy Future Academy
}

\begin{document}

\maketitle
\begingroup
\renewcommand{\thefootnote}{\textdagger}
\footnotetext{Corresponding author.}
\endgroup

\begin{abstract}
Current Large Multimodal Models (LMMs) struggle with spatial reasoning
tasks requiring viewpoint-dependent understanding, largely because they
are confined to a single, static observation. We propose
\textit{Thinking with Novel Views} (TwNV), a paradigm that integrates
generative novel-view synthesis into the reasoning loop: a Reasoner LMM
identifies spatial ambiguity, instructs a Painter to synthesize an
alternative viewpoint, and re-examines the scene with the additional
evidence. Through systematic experiments we address three research
questions. \textbf{(1)~Instruction format:} numerical camera-pose
specifications yield more reliable view control than free-form language.
\textbf{(2)~Generation fidelity:} synthesized view quality is tightly
coupled with downstream spatial accuracy.
\textbf{(3)~Inference-time visual scaling:} iterative multi-turn view
refinement further improves performance, echoing recent scaling trends
in language reasoning. Across four spatial subtask categories and four
LMM architectures (both closed- and open-source), TwNV consistently
improves accuracy by +1.3 to +3.9\,pp, with the largest gains on
viewpoint-sensitive subtasks. These results establish novel-view
generation as a practical lever for advancing spatial intelligence of LMMs.
\end{abstract}

\section{Introduction}
\label{sec:intro}
\input{sections/introduction}

\section{Related Work}
\subsection{MLLM}

The rapid advancement of MLLMs, such as GPT-4o~\cite{gpt4},
Gemini~\cite{gemini2.5}, and LLaVA~\cite{llava}, has set new benchmarks
in general-purpose visual perception and semantic understanding. By
leveraging large-scale vision-language pre-training, these models exhibit
remarkable zero-shot capabilities in descriptive tasks. However, their
``spatial intelligence'', the ability to comprehend the dynamic
world, remains a significant bottleneck. To address this, existing
research has branched into two directions. Architectural approaches,
including Spatial-MLLM~\cite{spatialmllm}, VLM-3R~\cite{vlm3r}, and
3DThinker~\cite{3dthinker}, introduce explicit geometric biases or 3D
representations into traditional MLLMs. SpatialBot~\cite{spatialbot}
augments visual encoders with depth, while VILASR~\cite{VILASR} enables
visual chain-of-thought by drawing auxiliary annotations on images. Data
scaling efforts, such as SpatialVLM~\cite{spatialvlm},
SpatialRGPT~\cite{spatialrgpt}, VST~\cite{vst}, and
SenseNova-SI~\cite{cai2025scaling}, expand spatial-specific supervision
through synthetic or densely annotated datasets. Despite this, most
methods rely on static supervised fine-tuning, understanding the world
solely from a single view. We instead propose to think with generative
novel views.

\subsection{Generative-Augmented Understanding}

Rooted in Analysis-by-Synthesis, Generative-Augmented Understanding links visual synthesis to structural world understanding. One line uses novel views as a reasoning medium: Chain-of-View Prompting~\cite{zhao2026cov} selects informative views from pre-captured 3D scenes for embodied QA, and Think3D~\cite{zhang2026think3d} uses generative 3D reconstruction for interactive viewpoint manipulation; related work~\cite{liu2025abstract, chen2025geometrically} argues that 3D-space perspective manipulation is essential for spatial reasoning. A complementary line builds the generative side: Thinking with Camera~\cite{liao2025thinking} treats camera parameters as language to bridge camera understanding and generation, PreciseCam~\cite{bernal2025precisecam} and DualCamCtrl~\cite{zhang2025dualcamctrl} improve the geometric fidelity of camera-controlled synthesis, and Depth Anything 3~\cite{depth3} recovers precise geometry from arbitrary views. Our \textit{Thinking with Novel Views} paradigm builds on both: unlike CoV, we synthesize novel views on-the-fly rather than rely on pre-captured ones; unlike camera-control methods focused on generation quality alone, we close the loop by using generated views as a reasoning medium and iteratively refining them through verification feedback.

\input{sections/3-setup}

\input{sections/4-dis}

\input{sections/5-con}

\bibliographystyle{plainnat}
\bibliography{main}

\input{sections/supp}

\end{document}

%% file: sections/introduction.tex
Current Large Multi-modal Models (LMMs) have traditionally operated under a ``Single-Pass Perception'' paradigm, where reasoning is strictly conditioned on the static content of the provided input image~\cite{llava, gpt4, internvl}. Recently, several works have attempted to evolve this toward a broader ``Thinking with Images'' (TwI) framework by enabling models to invoke external tools for image manipulation, such as cropping, zooming, or 2D rotation~\cite{gpt4tools, vst}. However, our preliminary investigation reveals a critical bottleneck: while these heuristic 2D transformations may improve local feature recognition, they offer negligible or even negative gains in complex spatial reasoning. Concretely, when we equip GPT-5~\cite{openai2025gpt5} with a tool-based grounding pipeline that lets the model identify and crop task-relevant regions, overall spatial reasoning accuracy \emph{drops} by 0.8 percentage points relative to the single-image baseline (69.6\% vs.\ 70.4\%), with the deficit widening to 2.0\,pp on multi-object relationship questions. This finding suggests that re-framing within the original 2D plane is fundamentally insufficient for spatial cognition: cropping and zooming cannot reveal occluded geometry or disambiguate 3D relational structures that are simply invisible from the input viewpoint. Humans, by contrast, resolve such ambiguities through \emph{mental simulation} --- we do not merely re-orient what we already see, but mentally construct entirely new perspectives to build a more complete spatial workspace.

A parallel line of work injects 3D structure into VLMs via depth maps, point clouds, or features from explicit 3D reconstruction~\cite{spatialvlm, vlm3r, spatialbot, zhang2026think3d}. However, recovering 3D geometry from a single image is fundamentally ill-posed and scale-ambiguous~\cite{zeng2024rsa}, so the geometry these methods inject is intrinsically noisy in the single-image regime that most spatial-reasoning benchmarks present. They also discard the image priors (textures, semantic context, fine-grained appearance) that pretrained VLMs are trained to read. Generative novel-view synthesis avoids both trade-offs: pretrained image generators have absorbed 3D-aware, world-model-level visual priors during web-scale pretraining~\cite{gabeur2026imagegenerators} and can be repurposed to render the same scene from a different camera pose~\cite{liu2023zero123}, producing a new RGB image that keeps the VLM within its trained input distribution while supplying geometric evidence the original view lacks.

\begin{figure}[!t]
    \vspace{-15pt}
    \centering
    \includegraphics[width=\linewidth]{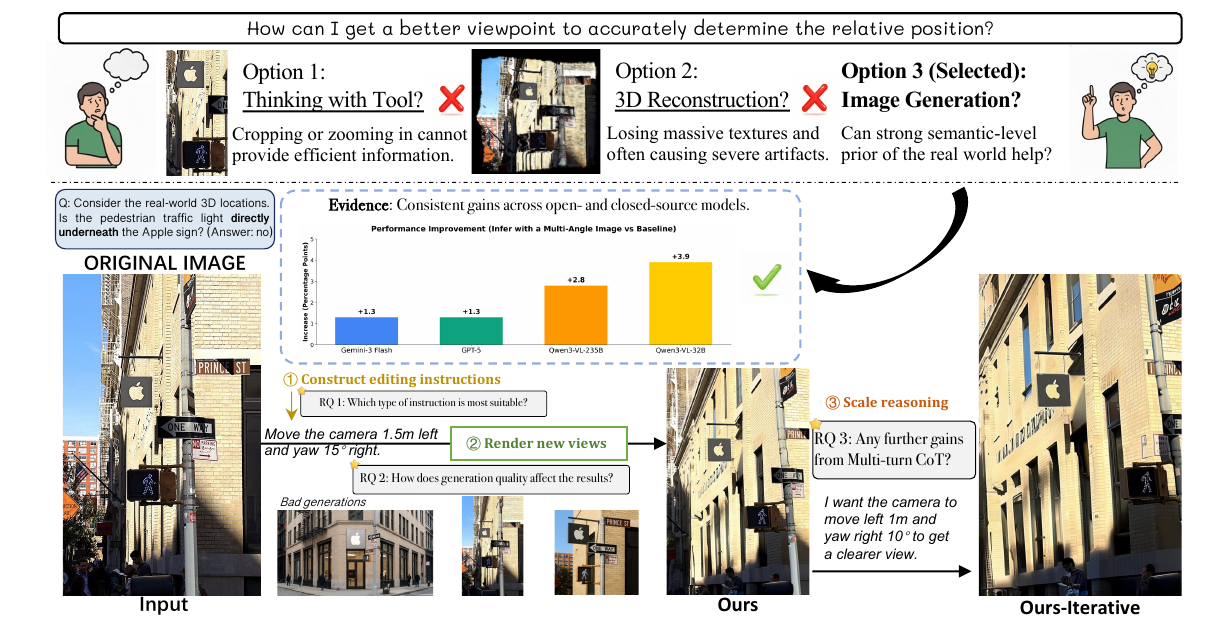}
    \caption{\textbf{Overview of the \textit{Thinking with Novel Views} paradigm.}
    \textit{(Top)}~Among three strategies for resolving viewpoint ambiguity,
    only generative novel-view synthesis provides sufficient 3D information
    while preserving semantics.
    \textit{(Center)}~Consistent accuracy gains (+1.3 to +3.9\,pp) across
    closed- and open-source VLMs.
    \textit{(Bottom)}~The system \textbf{(1)}~constructs a camera-motion
    instruction (\textbf{RQ\,1}), \textbf{(2)}~renders a novel view
    (\textbf{RQ\,2}), and optionally \textbf{(3)}~refines iteratively
    (\textbf{RQ\,3}). Details in \S\ref{sec:discussion}.}
    \label{fig:teaser}
    \vspace{-15pt}
    \end{figure}

In this paper, we propose a systematic shift from 2D tool-calling to a generative \textit{Thinking with Novel Views} (TwNV) paradigm. Rather than treating image generation as a creative end-goal, we reconceptualize it as a dynamic, 3D-aware reasoning workspace. As illustrated in Fig.~\ref{fig:teaser}, we introduce a three-stage pipeline: an LMM first acts as the \textbf{Planner}, proactively determining which novel camera pose would best disambiguate the spatial query; a generative model then serves as the \textbf{Synthesizer}, rendering the requested viewpoint by simulating ego-motion such as translation, panning, or tilting; finally, the LMM acts as the \textbf{Reasoner}, jointly interpreting the original and synthesized images to produce an answer. By transforming a static single-image query into an iterative, generation-augmented reasoning process that dynamically explores the 3D scene, we move beyond the perspective constraints of the initial input. As the evidence in Fig.~\ref{fig:teaser} shows, this approach yields consistent accuracy gains (+1.3 to +3.9 percentage points) across four VLMs spanning frontier closed-source systems and open-source models of varying scale, substantially outperforming the tool-based grounding baseline (e.g., 71.7\% vs.\ 69.6\% overall on GPT-5, with a striking 5.6\,pp advantage on multi-object relationships).

Building on this foundation, we conduct an in-depth exploration into the
governing principles of the TwNV paradigm, structured around five critical
dimensions:

\textbf{Scene Coverage.} Performance gains are non-uniform across spatial
subtasks: viewpoint-sensitive categories (orientation, multi-object
relationships) benefit the most, while size estimation sees slight
degradation, likely because novel views alter apparent object scale.
Detailed analysis is in \S\ref{sec:discussion}.

\textbf{Model Robustness.} We evaluate TwNV across a diverse spectrum of
model capacities, from Gemini-3-Flash~\cite{gemini2025} and GPT-5~\cite{openai2025gpt5} to the
open-source Qwen3-VL series (235B and 32B parameters)~\cite{bai2025qwen3vl}. The
paradigm delivers consistent improvements at every scale, with a notable
\textit{``Small-Model Dividend''}: the relative gain is more pronounced in
smaller models (e.g., +3.9\,pp for Qwen3-VL-32B vs.\ +1.3\,pp for
GPT-5), suggesting that explicit view synthesis serves as a compensatory
mechanism for parameter-constrained models by offloading 3D reasoning to
an external visual workspace.

\textbf{Instruction Design (\textbf{RQ\,1}).} To maximize the efficacy of the TwNV framework, we assess the optimal instruction interface for the Synthesizer. Our results indicate that continuous 3D camera parameters (\eg, explicit translation offsets and rotation angles) vastly outperform both free-form natural language descriptions and discrete categorical instructions. We hypothesize that providing explicit geometric priors minimizes the ``semantic noise'' inherent in linguistic prompts, thereby ensuring the generation of geometrically precise viewpoints that serve as faithful cues for downstream reasoning.

\textbf{Generation Fidelity (\textbf{RQ\,2}).} We examine how the quality of the Synthesizer influences the final reasoning outcome. By training a specialized pose-aware novel-view editing model to ensure geometric precision, we substantiate the hypothesis that a \emph{better Synthesizer} indeed yields a \emph{better Reasoner}. Our analysis reveals a positive correlation between the accuracy of viewpoint synthesis and the success rate of subsequent spatial inference, underscoring that the fidelity of the visual workspace is a decisive factor in the robustness of the entire pipeline.

\textbf{Inference-Time Visual Scaling (RQ\,3).} Finally, we ask whether
visual reasoning can be scaled at inference time, analogous to
Chain-of-Thought prompting in the textual domain. We demonstrate that
iterative refinement scales reasoning performance effectively with
additional rounds. In each round, the Reasoner evaluates its previous
generation, the Planner updates camera instructions, and the Synthesizer
produces improved viewpoints. This iterative loop allows the system to
progressively converge on the most diagnostic visual evidence, showing
that increased inference-time computation in the visual domain translates
directly into superior spatial intelligence. Under a matched VLM-call
budget, our image-based scaling outperforms text-only self-reflection
by $+3.0$\,pp ($78.7\%$ vs.\ $75.7\%$ on Gemini-3-Flash). The same
compute spent on a new synthesized view yields a larger gain than
re-examining the original one.

Through this systematic analytical lens, our work operationalizes generative novel-view synthesis as a core reasoning mechanism for spatial understanding. By characterizing the interplay among instruction precision, generative fidelity, model capacity, and iterative refinement, we establish a principled framework for the \textit{Thinking with Novel Views} paradigm that delivers robust and consistent gains across diverse spatial reasoning tasks and model families.

%% file: sections/3-setup.tex
\section{Pipeline, Dataset, and Evaluation}
\label{sec:setup}

\begin{figure}[t]
    \vspace{-15pt}
    \centering
    \includegraphics[width=1.0\textwidth]{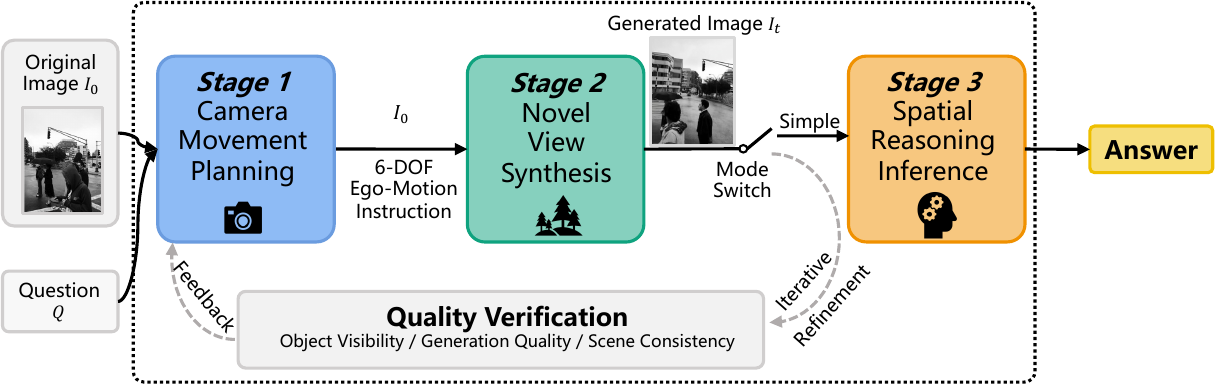}
    \caption{\textbf{The TwNV pipeline.} A Planner VLM proposes a 6-DOF camera-motion instruction, a Synthesizer renders the target view $I_t$, and a Reasoner VLM jointly interprets $\{I_0, I_t\}$. \textit{Iterative Mode} adds a Quality Verifier that rejects $I_t$ and feeds diagnostic feedback to the Planner for up to $N$ refinement rounds. 
    }
    \label{fig:pipeline}
    \vspace{-10pt}
\end{figure}

\subsection{Pipeline}
\label{sec:pipeline}

Our pipeline (Fig.~\ref{fig:pipeline}) transforms a single-image spatial query into a multi-perspective reasoning process through three stages and an optional iterative extension.

\noindent\textit{Stage~1: Camera Movement Planning.}
Given image $I_0$ and question $Q$, a VLM produces a 6-DOF ego-motion instruction $(\Delta x,\Delta y,\Delta z,\mathrm{yaw},\mathrm{pitch},\mathrm{roll})$ as geometric guidance for the synthesizer (\S\ref{sec:instruction}).

\textit{Stage~2: Novel View Synthesis.}
A generative model conditioned on $I_0$ and the instruction outputs a synthesized image $I_t$ from the target perspective (\S\ref{sec:generation}).

\textit{Stage~3: Spatial Reasoning.}
In \textbf{Simple Mode} ($N{=}0$), the VLM reasons over $\{I_0,I_t\}$. In \textbf{Iterative Mode} ($N{>}0$), a Quality Verifier evaluates $I_t$ on object visibility, generation quality, and scene consistency; rejected views trigger diagnostic feedback to the Planner for revised instructions. The VLM then reasons over $\{I_0,I^*\}$, where $I^*$ is the best accepted view (falling back to $I_0$ if all are rejected). This inference-time visual scaling strategy is analyzed in \S\ref{sec:capacity}.

\subsection{Evaluation Benchmark}
\label{sec:dataset}

\begin{figure}[t]
  \vspace{-15pt}
  \centering
  \includegraphics[width=0.7\textwidth]{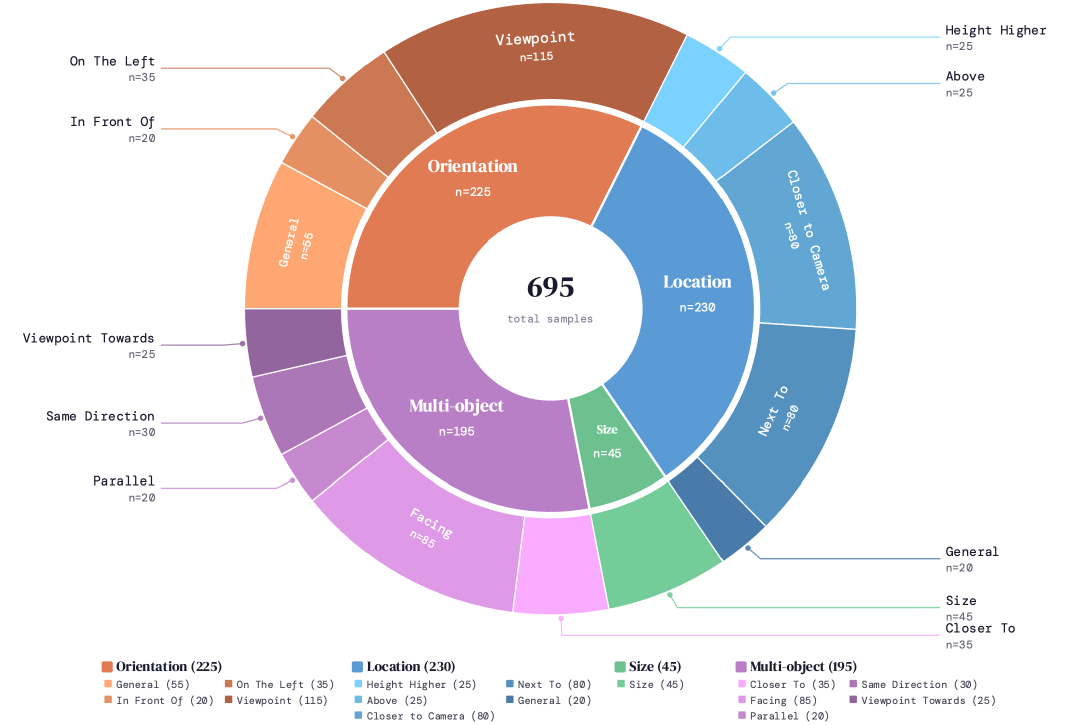}
  \caption{Benchmark distribution: 695 samples across 4 categories and 15 subcategories.}
  \label{fig:dataset}
  \vspace{-15pt}
\end{figure}

We assemble \textbf{695 samples} from two sources: \textbf{575} from 3DSRBench~\cite{ma2025_3dsrbench} via stratified sampling across all 12 subcategories, and \textbf{120} from RealWorldQA~\cite{xai2024realworldqa} focusing on spatially relevant subcategories (orientation, size, position). All samples are organized into four categories (Fig.~\ref{fig:dataset}), each probing 3D properties inherently ambiguous under single-view projection:

\textbf{Orientation} ($n{=}225$): directional pose including viewpoint, heading, and egocentric left/right and front/behind judgments.
\textbf{Location} ($n{=}230$): spatial position reasoning including depth ordering, vertical comparison, proximity, and relative layout.
\textbf{Size} ($n{=}45$): real-world physical dimension comparisons.
\textbf{Multi-Object} ($n{=}195$): relational reasoning across entities, including facing direction, relative proximity, co-orientation, and alignment.

\subsection{Implementation Details}
\label{sec:impl}

\textit{Reasoning Models.}
Closed-source: \textbf{Gemini-3-Flash}~\cite{gemini2025} and \textbf{GPT-5}~\cite{openai2025gpt5}; open-weight: \textbf{Qwen3-VL-235B} and \textbf{Qwen3-VL-32B}~\cite{bai2025qwen3vl}. All four serve as Planner, Verifier, and Reasoner.

\textit{Synthesis Models.}
\textbf{GPT-Image-1}~\cite{openai2025gptimage}, \textbf{Nano Banana Pro}~\cite{nanabananapro2025}, \textbf{Qwen-Image-Edit}~\cite{wu2025qwenimage}, and \textbf{CamCraft} (\emph{Ours}), a pose-conditioned editing model for geometrically faithful viewpoint synthesis (details in Appendix).

\textit{Answer Judging.}
Deterministic string matching first; \textbf{Gemini-3-Flash} as semantic judge when ambiguous.

\textit{Conditions.}
\textbf{Baseline}: single-pass on $I_0$;
\textbf{Ours}: Simple Mode, reasoning from $\{I_0,I_t\}$;
\textbf{Ours-Iterative}: Iterative Mode, reasoning from $\{I_0,I^*\}$.
All conditions, including the Baseline, are reported as the majority vote of $K{=}3$ independent runs, so that observed differences reflect paradigm choice rather than sampling variance.

%% file: sections/4-dis.tex
\section{Discussion}
\label{sec:discussion}

Generative novel views can supply spatial evidence the original image lacks, but their usefulness depends on the scene, the reasoning model, the instruction format, and the generator. We address these factors in the five subsections below.

\subsection{Scene Coverage: Is Generated Image Universally Effective Across Scenes?}
\label{sec:scene}

To investigate the scope of applicability, we examine whether generated views provide consistent benefits across diverse scene distributions rather than being confined to ``generator-friendly'' scenarios. We evaluate our paradigm on both a general-purpose and the proposed spatial-oriented understanding benchmark, using \gpt{} as the reasoner and GPT-Image-1 as the painter \textit{without any task-specific optimizations}.

\begin{figure}[htbp]
    \vspace{-10pt}
    \centering
    \includegraphics[width=0.9\textwidth]{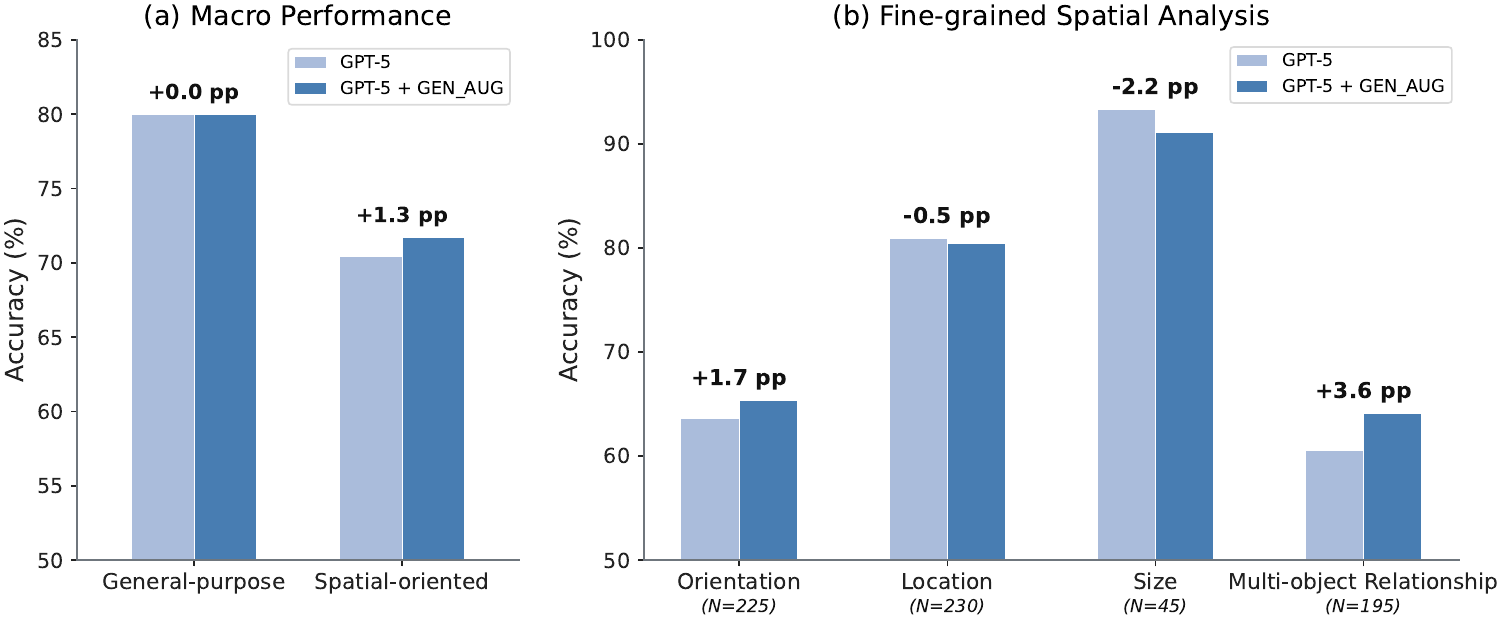}
\caption{\textbf{Effect of generative view augmentation.} (a)~Macro accuracy on general-purpose vs.\ spatial-oriented benchmarks (no change on general, $+1.3$pp on spatial). (b)~Breakdown by spatial subtask; sample size $N$ is annotated under each category, and values above bars show changes over the GPT-5 baseline. }
    \label{fig:scene_analysis}
    \vspace{-10pt}
\end{figure}

As expected, as illustrated in Figure~\ref{fig:scene_analysis}(a), our preliminary analysis on general tasks (encompassing basic perception, recognition, and subject-related QA) indicates that novel views offer negligible assistance, as the required information is already inherently sufficient within the primary perspective.

In contrast, for the 3D-oriented benchmark, Figure~\ref{fig:scene_analysis}(a) shows that dynamic view compensation improves the overall accuracy from 70.4\% to 71.7\%. A fine-grained analysis (Figure~\ref{fig:scene_analysis}(b)) reveals that our method yields obvious improvements in \textit{Orientation} (+1.7pp) and \textit{Multi-object Relationship} (+3.6pp) tasks. The significant lift in the latter suggests that novel perspectives are particularly effective for complex relational reasoning, where a single viewpoint may suffer from occlusion or perspective ambiguity. While the model maintains comparable performance in \textit{Location} (80.4\% vs. 80.9\%), we observe a minor degradation in the \textit{Size} category. We hypothesize that while generative synthesis excels at providing structural and relational cues, it may also introduce subtle metric inconsistencies or generative artifacts that act as noise, thereby impacting fine-grained scale estimation. Nevertheless, the consistent gains in higher-level reasoning tasks underscore the profound potential of incorporating novel views to bolster spatial intelligence.

Errors fall into three categories: \textit{wrong instruction}, where the prompt under- or mis-specifies the requested view; \textit{bad generation}, where the synthesized image has geometric errors, artifacts, or missing objects; and \textit{VL failure}, where the reasoner errs even when the view is correct. Fig.~\ref{fig:failure_and_avg}(a) shows that bad generation is the main bottleneck, which is why we examine instruction quality in Section~\ref{sec:instruction} and the generator in Section~\ref{sec:generation}.

\subsection{Model Robustness: Only High-Capacity Reasoners or Also Weaker Ones?}
\label{sec:model}

A practical concern is whether our paradigm only amplifies
already-strong models or also benefits weaker reasoners. We evaluate the
same generation protocol across four understanding backbones of varying
capacity. For each backbone, the \textit{baseline} uses the original
single-view input, while \gemodels{} augments it with our novel-view
procedure under identical settings.

\begin{figure}[t]
    \vspace{-10pt}
    \centering
    \includegraphics[width=\textwidth]{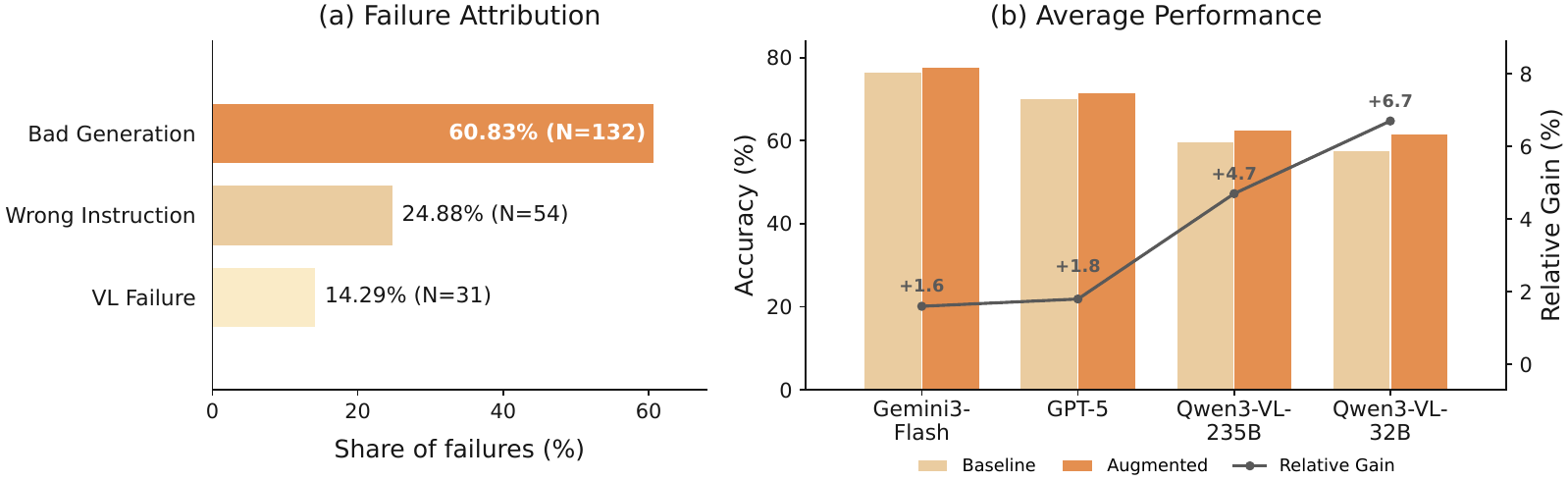}
    \caption{\textbf{Failure attribution and per-backbone gains.}
    (a)~Error sources on \gpt{}+\gptImage{}: bad generation dominates (60.8\%), followed by wrong instructions (24.9\%) and VL failures (14.3\%).
    (b)~Spatial-oriented accuracy for four backbones: baseline (single view) vs.\ \gemodels{}-augmented (bars), with relative gain $(\text{aug}{-}\text{base})/\text{base}$ on the right axis (line). Weaker backbones gain more, peaking at +6.7\% for \qwenMiddle{}.}
    \label{fig:failure_and_avg}
    \vspace{-15pt}
\end{figure}

Fig.~\ref{fig:failure_and_avg}(b) shows two things. The paradigm helps every backbone we tested, and the smaller the model, the larger its relative gain: \qwenMiddle{} improves the most at $+6.7$\%, while the stronger \geminiFlash{} improves the least. We call this the \textit{Small-Model Dividend}. Smaller models have weaker internal 3D representations, so an explicit novel view gives them more spatial information than they could recover on their own.

\subsection{Prompting Strategies: Are Explicit 3D Camera Parameters Better than Descriptive Motion Language?}

\label{sec:instruction}

\begin{figure}[h]
    \vspace{-15pt}
    \centering
    \includegraphics[width=1.0\textwidth]{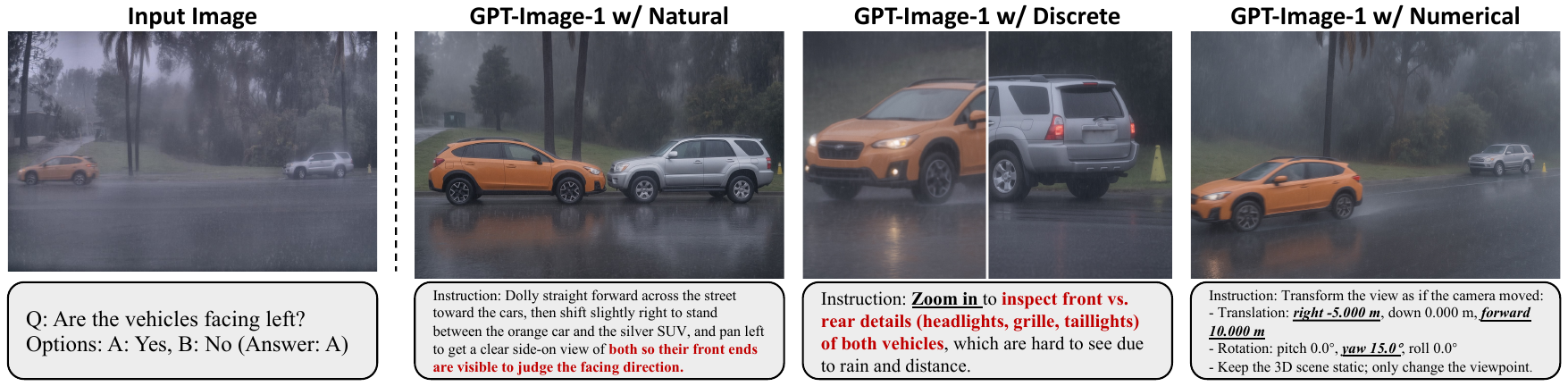}
    \caption{Qualitative comparison of three instruction paradigms.
    \textbf{Natural}: free-form motion language.
    \textbf{Discrete}: categorical directives
    (\emph{e.g.}, ``zoom in'', ``side view'').
    \textbf{Numerical}: explicit 6-DoF pose changes.
    Natural and discrete prompts introduce redundant semantic cues
    that cause spatial hallucinations (e.g., shifted object positions),
    whereas numerical instructions preserve the original scene layout.}
    \label{fig:instruction_type}
    \vspace{-5pt}
\end{figure}

The efficacy of generative-augmented reasoning is highly contingent on
instruction quality. We therefore investigate which format best elicits
informative and reliable viewpoint changes. Specifically, we compare
three paradigms: descriptive natural language, discretized view changes,
and precise numerical camera pose changes, as visualized in
Figure~\ref{fig:instruction_type}.

We use end-to-end spatial-reasoning accuracy as a proxy for instruction quality. We pair two reasoners, \gpt{} and Gemini-3-Flash, with two painters, GPT-Image-1 and Nano Banana Pro, so that prompt-engineering quirks or model-specific preferences do not drive the result.

\begin{figure}[h]
    \vspace{-10pt}
    \centering
    \includegraphics[width=0.7\textwidth]{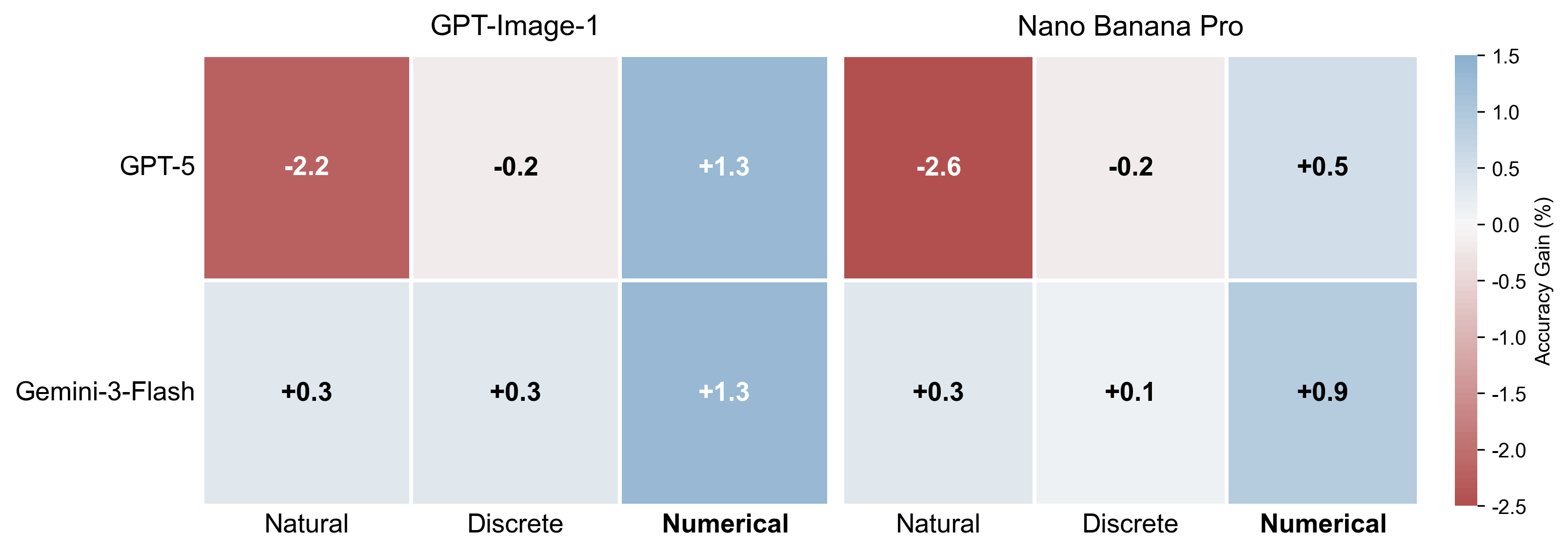}
    \caption{Accuracy gain (\%) over the single-view baseline for three
    instruction formats across two reasoners (rows) and two painters
    (column groups). \textcolor{blue}{Blue}: improvement;
    \textcolor{red}{red}: degradation.}
    
    \label{fig:instruction_heatmap}
    \vspace{-10pt}
\end{figure}

As illustrated in Figure~\ref{fig:instruction_heatmap}, numerical camera
pose changes yield the most robust performance gains, remaining
consistently positive across all four reasoner--painter combinations
($+0.5\%$ to $+1.3\%$). In contrast, natural language instructions
exhibit a strong reasoner-dependent effect: they produce marginal gains
with Gemini-3-Flash ($+0.3\%$) but substantially \emph{degrade}
performance with \gpt{} ($-2.2\%$ / $-2.6\%$). We attribute this
discrepancy to the semantically redundant cues inherent in free-form
descriptions (mentioning object identities, scene context, or intended
inspection goals), which prompt the editing model to hallucinate spatial
modifications beyond the intended viewpoint shift, such as altering
object positions or scales. Discretized instructions reduce this risk
but still carry categorical semantics that can be misinterpreted,
resulting in near-zero or marginal gains across most settings. Numerical
instructions, by conveying only geometric transformation parameters
(translation in meters, rotation in degrees), minimize such
hallucinations and ensure the generative module produces viewpoint
changes that faithfully preserve the original scene layout. This
consistency underscores the advantage of using explicit geometric
constraints to improve spatial reasoning performance.

\subsection{Generation Quality: Would a “Better Painter” Make a Better Thinker in 3D?}

\label{sec:generation}

Even with carefully designed prompts, the overall performance can still be
bottlenecked by the generator's ability to preserve 3D fidelity and multi-view
consistency. In this section, we investigate the impact of generation quality
from two angles: (i)~how do existing mainstream image editing models, both
open-source and commercial, affect the downstream spatial reasoning accuracy,
and (ii)~can a generator purpose-built for pose-conditioned camera
editing further close the gap? Concretely, we keep the understanding
model (\gpt) and task protocol fixed, and swap the image editor among four
candidates: two open-source models (\qwenImage{} and \nano{}), one commercial
model (\gptImage{}), and our own pose-conditioned editor, \joyomni{},
trained specifically on large-scale 3D data for geometrically faithful
view synthesis. Full architecture, training data, and training configuration for \joyomni{} are detailed in Appendix~\ref{sec:supp_nvs}.

\begin{table}[t]
\vspace{-15pt}
\centering
\small
\caption{Comparison of different image editing models in our pipeline.
\gpt is used as the fixed understanding model. The results show the accuracy (\%) 
for each sub-category and the final voting accuracy.}
\label{tab:generation_model_comparison}
\resizebox{0.95\linewidth}{!}{
\begin{tabular}{lccccc}
\toprule
\textbf{Category} & \textbf{w/o Gen} & \qwenImage{} & \nano{} & \gptImage{} & \joyomni{} \\
\midrule
Orientation & 63.6 & 61.8 & 60.9 & 65.3 & 65.3 \\
Location & 80.9 & 77.8 & 83.0 & 80.4 & 82.6 \\
Size & 93.3 & 86.7 & 91.1 & 91.1 & 91.1 \\
Multi-object & 60.5 & 61.5 & 63.6 & 64.1 & 66.2 \\
\midrule
Overall & 70.4 & 68.6 & 70.9 & 71.7 & 72.9 \\
\bottomrule
\end{tabular}
}

\end{table}

Table~\ref{tab:generation_model_comparison} reports the results. Not every
generator improves over the no-generation baseline: \qwenImage{} decreases the
overall accuracy by 1.8\,pp, suggesting that low-fidelity edits can introduce
misleading visual evidence that actively harms reasoning. \nano{} yields a
marginal gain (+0.5\,pp), while \gptImage{} provides a moderate improvement
(+1.3\,pp). \joyomni{} achieves the best overall accuracy (72.9\%), a +2.5\,pp
gain over the baseline, and leads on three of the four subtasks. The
progression from \qwenImage{} to \joyomni{} reveals a clear trend: the more
geometrically faithful the generated view, the greater the benefit for spatial
reasoning. In particular, a generator must reach a certain fidelity threshold
before its outputs help rather than hurt the downstream reasoner.

The improved generator also shifts where failures concentrate. Compared to \gptImage{}, the \emph{Bad Generation} share drops from 60.8\% to 45.2\%, while the \emph{Wrong Instruction} share rises from 24.9\% to 43.1\%. The instructions themselves did not get worse; cleaning up generation errors simply leaves instruction quality as the next bottleneck, which we address in \S\ref{sec:capacity}.

\subsection{Capacity Scaling: Can We Scale Reasoning with Image Sampling and Verification Like Text?}
\label{sec:capacity}

\begin{figure}[h]
    \vspace{-15pt}
    \centering
    \includegraphics[width=1.0\textwidth]{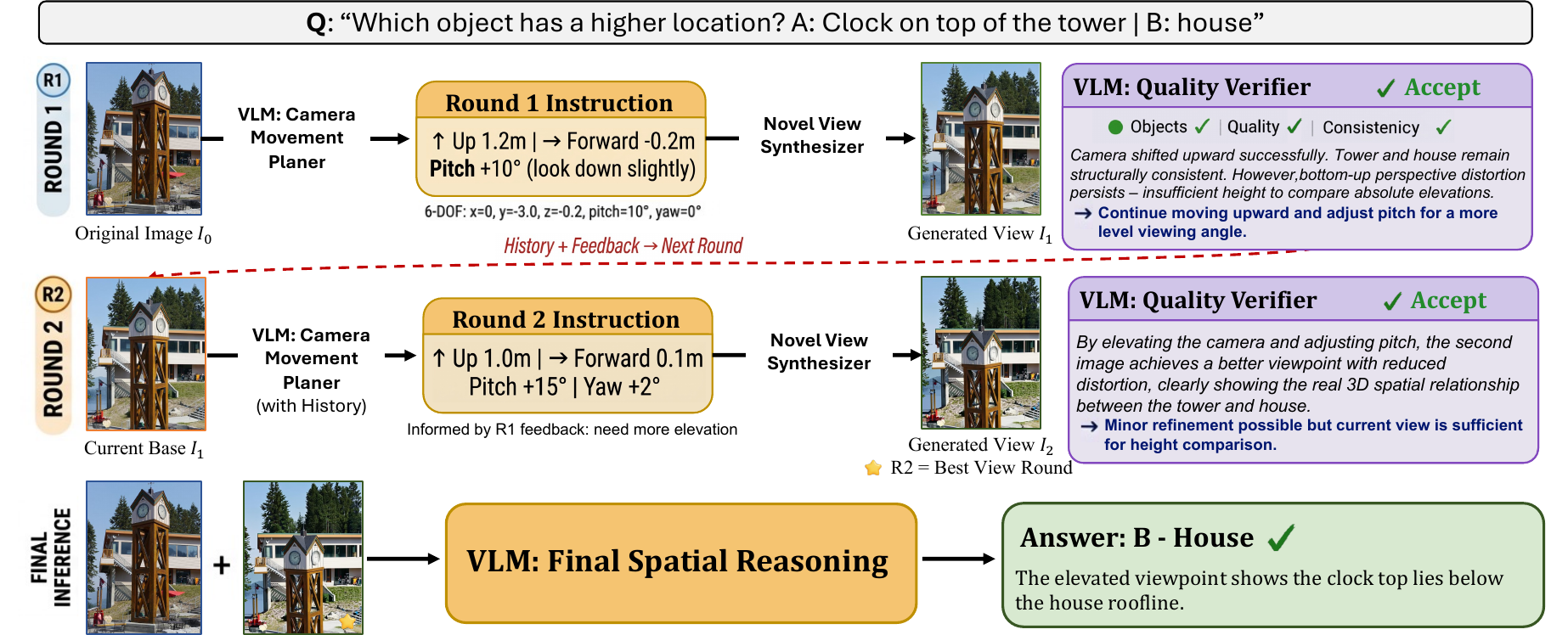}
    \caption{Iterative viewpoint refinement for height comparison: the VLM progressively elevates the camera across two rounds guided by verification feedback, producing a distortion-free view that enables correct spatial reasoning.}
    \label{fig:iterative}
    \vspace{-10pt}
\end{figure}

\begin{table}[t]
\vspace{-5pt}
\centering
\caption{Test-time scaling comparison. \textit{Top}: text-only self-reflection on the original image ($R$: number of reflection rounds). \textit{Bottom}: image-based iterative refinement ($N$: verification--refinement rounds; $N{=}0$: Simple Mode). Text $R{=}1$ and Image $N{=}0$ both use two VLM calls per run (matched VLM-call budget).}
\label{tab:scaling}
\resizebox{0.7\linewidth}{!}{
\begin{tabular}{ll ccccc}
\toprule
\textbf{Strategy} & \textbf{Config} & \textbf{Orient.} & \textbf{Loc.} & \textbf{Size} & \textbf{Multi-obj} & \textbf{Overall} \\
\midrule
\multirow{3}{*}{Text reflection}
 & Baseline ($R{=}0$)        & 68.9 & 82.6 & 93.3 & 74.9 & 76.7 \\
 & Best ($R{=}1$)            & 69.3 & 81.3 & 93.3 & 72.3 & 75.7 \\
 & Worst ($R{=}4$)           & 68.4 & 77.0 & 91.1 & 68.2 & 72.7 \\
\midrule
\multirow{3}{*}{Image iterative}
 & $N{=}0$ (Simple)          & 73.3 & 85.2 & 91.1 & 74.4 & 78.7 \\
 & $N{=}1$                   & 72.4 & \textbf{86.5} & 91.1 & \textbf{76.9} & \textbf{79.6} \\
 & $N{=}2$                   & 72.4 & 86.1 & 91.1 & 76.4 & 79.3 \\
\bottomrule
\end{tabular}
}
\vspace{-15pt}
\end{table}

Section~\ref{sec:generation} flagged \emph{Wrong Instruction} as the new bottleneck (43.1\%). Can we reduce these errors at test time by letting the model verify its own generated views and refine the camera instructions, much like textual self-consistency?

\textit{Image-based scaling via iterative refinement.}
We activate the Iterative Mode (\S\ref{sec:pipeline}) with
$N \in \{1, 2\}$ verification rounds on the best Simple-Mode configuration
(\geminiFlash{} +  \joyomni{}). At each round, the understanding model inspects
the generated view, diagnoses whether it provides useful spatial evidence, and
if not, issues a revised camera instruction for re-generation.
A single verification round ($N{=}1$) yields a consistent overall
improvement ($78.7\% \!\to\! 79.6\%$), with the largest gains on
Location ($85.2\% \!\to\! 86.5\%$) and Multi-Object Relationship
($74.4\% \!\to\! 76.9\%$), both categories where viewpoint quality is most
critical (Table~\ref{tab:scaling}, bottom). However, adding a second round ($N{=}2$) already causes slight
regression ($79.3\%$), and further iterations continue to erode gains.
We attribute this to error accumulation: each additional
generation--verification cycle risks introducing artifacts or drifting from
the original scene context.
Fig.~\ref{fig:iterative} illustrates a successful two-round refinement where
the verifier guides the planner to progressively elevate the camera,
producing a distortion-free view that resolves a height comparison.

\textit{Comparison with text-only self-reflection.}
We compare against \emph{self-reflection}, where the VLM critiques its initial answer and re-reasons over the original image for up to $R \in \{1, \dots, 10\}$ rounds. Self-reflection \emph{consistently degrades} performance, fluctuating between $72.7\%$ and $75.7\%$ versus the single-pass baseline of $76.7\%$ (Table~\ref{tab:scaling}, top): textual deliberation cannot compensate for missing visual evidence. Image-based scaling at $N{=}1$ ($79.6\%$) beats the best self-reflection variant by $+3.9$\,pp, and the advantage holds at matched VLM-call budget: Simple Mode ($N{=}0$) already reaches $78.7\%$ versus text reflection $R{=}1$'s $75.7\%$. The gap is not from extra compute but from reasoning over a new viewpoint instead of the same one.

\textit{Where do the gains come from?} Table~\ref{tab:error_loop} shows the loop mostly removes \emph{Wrong Instruction} errors: the count drops from 81 to 24, a 70\% reduction, so the model is self-correcting its camera planning. \emph{Bad Generation} stays roughly flat. The loop can change what we ask the generator to render but not how faithfully it renders that request, so bad generation now accounts for 71.1\% of the remaining errors. Generation fidelity is the next bottleneck.

\begin{wraptable}{r}{0.3\textwidth}
\vspace{-20pt}
\centering
\small
\caption{Error attribution before and after iterative refinement.}
\label{tab:error_loop}
\resizebox{1.0\linewidth}{!}{
\begin{tabular}{l cc cc}
\toprule
& \multicolumn{2}{c}{\textbf{Simple}}
& \multicolumn{2}{c}{\textbf{Iter.}} \\
\cmidrule(lr){2-3} \cmidrule(lr){4-5}
\textbf{Error Type} & \# & \% & \# & \% \\
\midrule
Wrong Instr. & 81 & 43.1 & 24 & 16.9 \\
Bad Gen.     & 85 & 45.2 & 101 & 71.1 \\
VL Failure   & 22 & 11.7 & 17 & 12.0 \\
\midrule
\textbf{Total} & 188 & 100 & 142 & 100 \\
\bottomrule
\end{tabular}
}
\vspace{-15pt}
\end{wraptable}

These results paint a coherent picture across
Sections~\ref{sec:generation}--\ref{sec:capacity}: improving the generator
(\S\ref{sec:generation}) shifted the bottleneck from \emph{Bad Generation}
to \emph{Wrong Instruction}; iterative verification (\S\ref{sec:capacity})
now resolves most instruction errors, shifting the bottleneck back to
generation fidelity. Further progress will likely require advances on
\emph{both} fronts simultaneously (higher-fidelity 3D-aware generators
and more precise camera planning) to push beyond the current accuracy
ceiling.

%% file: sections/5-con.tex
\section{Conclusion}
\label{sec:conclusion}

We presented a systematic study of using image generation as a cognitive tool to augment spatial reasoning in VLMs. Rather than relying on a single, ambiguous 2D observation, our \emph{Thinking with Novel Views} (TwNV) pipeline forms a closed loop: a Planner proposes informative camera motions, a Synthesizer renders the requested viewpoint, a Reasoner interprets the enriched evidence, and a Quality Verifier feeds diagnostic signals back to the Planner for iterative refinement.

Across instruction design, generator capacity, and test-time scaling, three findings emerge. \textbf{(1)~Instruction matters:} structured pose-conditioned camera transformations reduce ambiguity over free-form view synthesis (\S\ref{sec:instruction}). \textbf{(2)~Better painter, better thinker:} a purpose-built generator (\joyomni{}) outperforms open-source and commercial alternatives, confirming that 3D-faithful generation is the key enabler of downstream gains (\S\ref{sec:generation}). \textbf{(3)~Visual scaling trumps verbal scaling:} iterative view synthesis improves accuracy and outperforms textual self-reflection (\S\ref{sec:capacity}). Error attribution reveals a two-stage bottleneck shift: improving the generator moves the dominant error from \emph{Bad Generation} to \emph{Wrong Instruction}, and iterative verification resolves most instruction errors, returning the bottleneck to generation fidelity, a diagnostic loop that charts the next round of improvement.

\textit{Limitations and future work.} Iteration shifts the residual bottleneck back to generation fidelity, where \joyomni{} still produces artifacts under large viewpoint changes. Stronger 3D-aware generators or a dedicated \emph{spatial VLM} fine-tuned on pose-conditioned planning data would each chip away at this gap, but neither addresses the deeper issue: planning, rendering, and verification are still handled by separate networks whose hand-offs accumulate disagreement and cause the verification loop to saturate after one round. A \emph{unified multimodal model} for understanding and generation~\citep{liao2025thinking, showo, januspro}, in which one network plans, renders, and verifies viewpoints, would replace this brittle pipeline. We see this as the most promising path toward stronger generative spatial intelligence. Our evaluation is also limited to static, rigid scenes; dynamic objects and predicates such as containment, support, and occlusion remain open. We hope this work establishes image generation as a principled and complementary axis, alongside textual chain-of-thought and tool use, for enhancing the spatial intelligence of foundation models.

%% file: sections/supp.tex
\clearpage
\setcounter{section}{0}
\setcounter{figure}{0}
\setcounter{table}{0}
\renewcommand{\thesection}{\Alph{section}}
\renewcommand{\theHsection}{supp.\Alph{section}}
\renewcommand{\thefigure}{\thesection.\arabic{figure}}
\renewcommand{\thetable}{\thesection.\arabic{table}}
\numberwithin{table}{section}
\numberwithin{figure}{section}

\begin{center}
  \LARGE\textbf{Supplementary Material}
\end{center}
\vspace{6pt}

\section{Novel View Synthesis Model: Training and Evaluation}
\label{sec:supp_nvs}

This appendix provides comprehensive details on the architecture, training data, data processing pipeline, training procedure, and dedicated evaluation of \joyomni{}, the novel view synthesis (NVS) model employed in Stage~2 of our pipeline (cf.\ Section~\ref{sec:pipeline} in the main paper).

\subsection{Architecture}
\label{sec:supp_arch}

\joyomni{} is built upon a 16B-parameter MMDiT backbone paired with a Qwen3-VL-8B-Instruct semantic encoder that encodes the text-form camera-pose instruction.  The reference (source) image is independently encoded by the Wan2.1 VAE into a latent representation, which is then concatenated channel-wise with the generation noise latent.  Joint self-attention is applied over both the reference and generation representations, allowing the model to attend to fine-grained appearance details while following the geometric transformation specified by the 6-DOF pose instruction $(\Delta x,\Delta y,\Delta z,\text{yaw},\text{pitch},\text{roll})$.  This design prioritises geometric consistency under explicit camera-pose control, directly targeting the dominant failure mode (\emph{Bad Generation}) identified in Section~\ref{sec:scene} of the main paper.

\subsection{Training Data}
\label{sec:supp_training_data}

\begin{figure}[t]
  \centering
  \includegraphics[width=\linewidth]{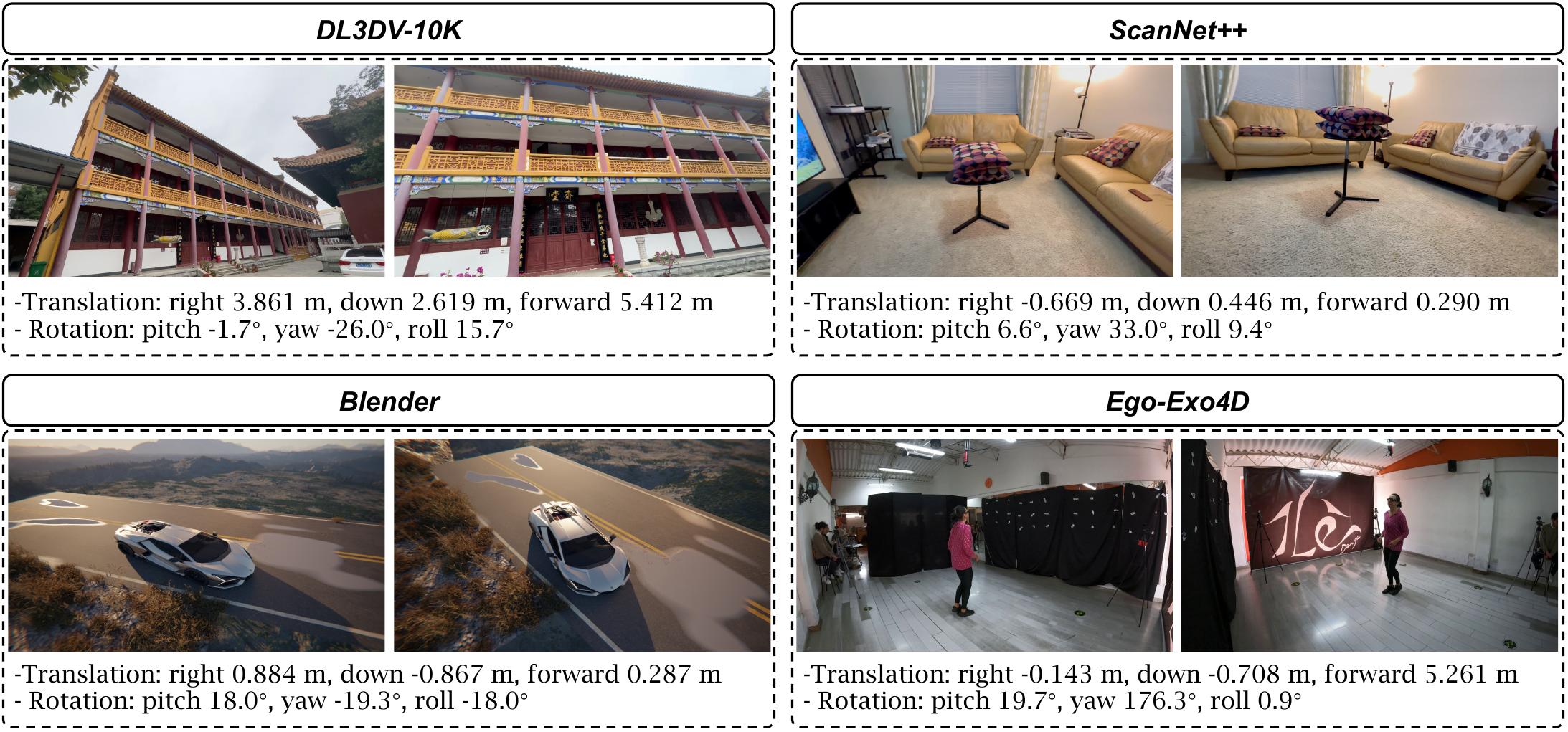}
  \caption{\textbf{Training data overview.}  Representative source--target image pairs from the four datasets used to train \joyomni{}.  Each panel shows a pair of images together with the annotated 6-DOF camera transformation (translation in metres; rotation in degrees).  The training set spans outdoor architecture (\emph{DL3DV-10K}), indoor rooms (\emph{ScanNet++}), photorealistic synthetic environments (\emph{Blender}), and human-centric multi-view captures (\emph{Ego-Exo4D}), collectively covering a wide range of scene types and camera-motion magnitudes.}
  \label{fig:supp_training_data}
\end{figure}

The model is trained on \textbf{7.6\,M image pairs} collected from four complementary 3D data sources, each contributing distinct scene characteristics and camera-motion distributions (see Figure~\ref{fig:supp_training_data} for representative examples):

\begin{itemize}[leftmargin=*,itemsep=2pt]
  \item \textbf{DL3DV-10K}~\cite{ling2024dl3dv}: Large-scale real-world captures spanning diverse indoor and outdoor environments with wide-baseline camera trajectories and substantial translation magnitudes.
  \item \textbf{ScanNet++}~\cite{yeshwanth2023scannetpp}: High-fidelity indoor reconstructions of room-scale scenes, providing dense viewpoint coverage with moderate camera motions.
  \item \textbf{Blender}: Synthetic scenes rendered with \emph{exact} ground-truth camera parameters, offering pixel-perfect pose supervision and diverse viewpoint transformations including large rotations.
  \item \textbf{Ego-Exo4D}~\cite{grauman2024egoexo4d}: Human-centric scenes captured from both egocentric and exocentric viewpoints, introducing large camera translations/rotations and diverse human activities.
\end{itemize}

\noindent
The combination of these sources ensures broad coverage of indoor/outdoor settings, synthetic/real imagery, human-centric scenarios, and camera motions ranging from subtle to extreme.

\subsection{Data Processing Pipeline}
\label{sec:supp_data_processing}

Two key processing steps are applied uniformly across all open-source datasets (DL3DV-10K, ScanNet++, and Ego-Exo4D; the Blender data already provides exact metric-scale poses and does not require either step).

\paragraph{ORB-based overlap filtering.}
For each candidate image pair, we compute ORB feature descriptors and perform brute-force matching.  Pairs with \emph{fewer than 10 verified matches} are discarded.  This filtering removes pairs whose views share negligible visual overlap (\eg, due to extreme camera baselines or entirely different rooms), which would otherwise introduce noisy supervision and destabilise training.

\paragraph{Metric-scale normalisation.}
Different datasets define camera poses in heterogeneous coordinate systems whose unit lengths do not correspond to physical metres.  To unify translation magnitudes across all sources, we leverage Depth Anything~3~\cite{lin2025depthanything3} to predict monocular metric-depth maps for both images in each pair.  We then back-project the known camera-relative depth values into 3D point clouds, and compute a least-squares scale factor by aligning the reconstructed point-cloud depths to the predicted metric depths.  The resulting factor is used to rescale the translation vector $(\Delta x, \Delta y, \Delta z)$ to \emph{metres}; rotations are intrinsically scale-invariant and therefore remain unchanged.  This procedure ensures that the model receives geometrically consistent, real-world-scale translation supervision regardless of the original dataset's coordinate conventions.

\subsection{Training Configuration}
\label{sec:supp_training_config}

We adopt a bucket-based resolution strategy with a base size of $1024$ to accommodate varying aspect ratios of the input images; each image pair is assigned to the resolution bucket that best preserves its native aspect ratio while keeping the total pixel count close to $1024\!\times\!1024$.  Each training step uses a global batch size of 2{,}048 samples.  To preserve the base model's text-to-image (T2I) generation capability and mitigate catastrophic forgetting, we adopt a mixed training strategy: 50\% of each batch consists of standard T2I data, while the remaining 50\% consists of camera-pose-conditioned editing pairs.  This yields an effective batch size of 1{,}024 editing pairs per step; consequently, 8k steps correspond to approximately one full epoch over the 7.6\,M editing pairs.

The model is initialised from the pre-trained T2I weights and fine-tuned end-to-end on the mixed data described above.  The convergence behaviour under this setting is analysed in Section~\ref{sec:supp_convergence}.

\subsection{Camera Pose Control Benchmark}
\label{sec:supp_benchmark}

\begin{figure}[t]
  \centering
  \includegraphics[width=\linewidth]{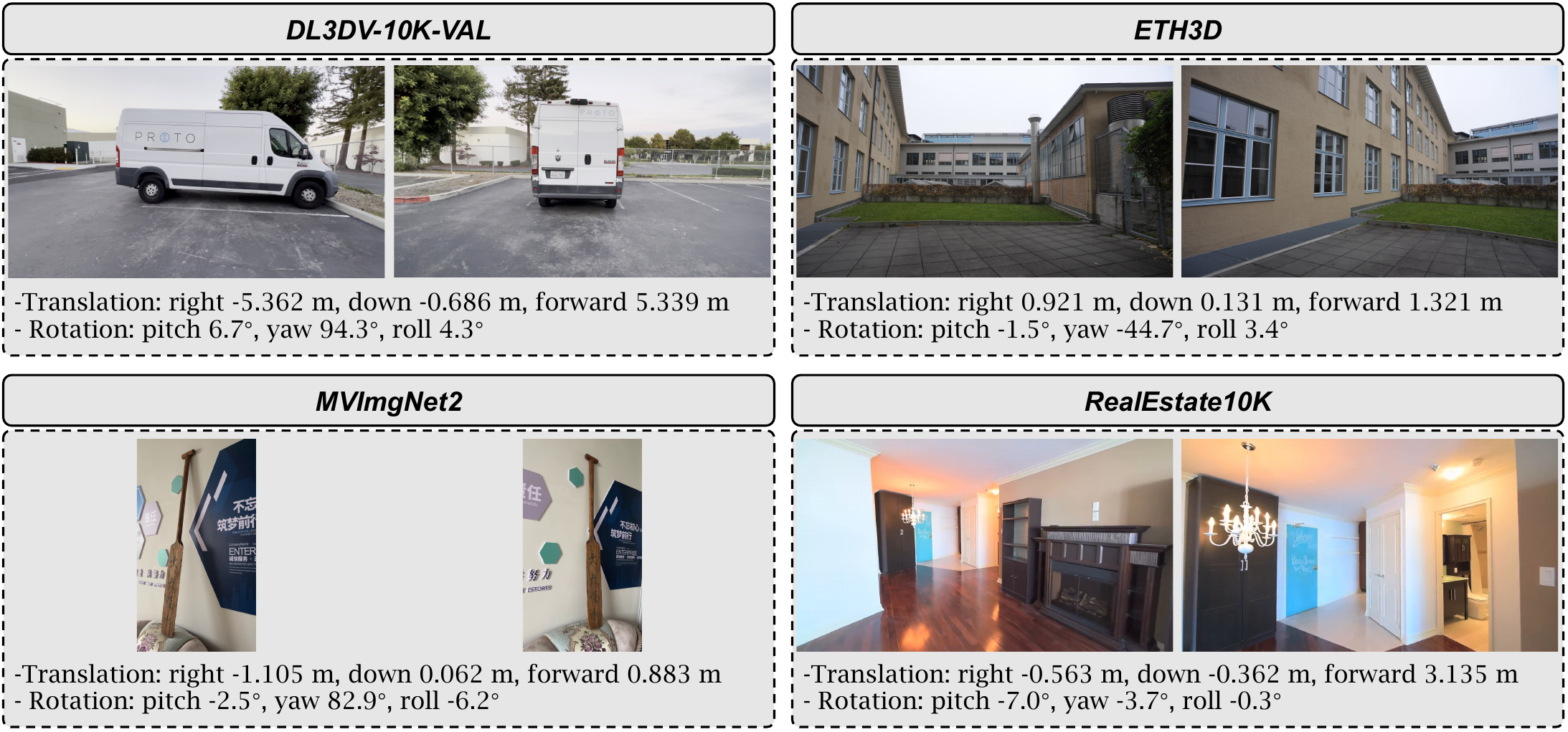}
  \caption{\textbf{Camera pose control evaluation benchmark.}  Representative source--target pairs from the four evaluation datasets with annotated 6-DOF transformations.  \emph{DL3DV-10K-VAL} serves as an in-domain test set (including manually curated challenging viewpoint transitions), while \emph{ETH3D}, \emph{MVImgNet2}, and \emph{RealEstate10K} provide out-of-domain evaluation across diverse scene types and camera-motion patterns.}
  \label{fig:supp_eval_benchmark}
\end{figure}

To rigorously evaluate the geometric fidelity of \joyomni{}, we construct a dedicated \textbf{camera pose control benchmark} comprising \textbf{168 evaluation pairs} that span both in-domain and out-of-domain data sources.  Figure~\ref{fig:supp_eval_benchmark} shows representative evaluation pairs.

\paragraph{Benchmark composition.}
Evaluation pairs are drawn from four datasets:

\begin{itemize}[leftmargin=*,itemsep=2pt]
  \item \textbf{DL3DV-10K-VAL} (\emph{in-domain})~\cite{ling2024dl3dv}: A held-out validation split of DL3DV-10K.  Samples include both \emph{randomly selected} pairs and \emph{manually curated} cases with semantically meaningful viewpoint changes (\eg, a side-view to front-view transition of the main subject).
  \item \textbf{ETH3D} (\emph{out-of-domain})~\cite{schoeps2017eth3d}: Outdoor multi-view stereo scenes featuring architectural structures and building fa\c{c}ades.
  \item \textbf{MVImgNet2} (\emph{out-of-domain})~\cite{han2024mvimgnet2}: Object-centric multi-view captures with moderate camera motions around everyday objects.
  \item \textbf{RealEstate10K} (\emph{out-of-domain})~\cite{zhou2018realestate10k}: Indoor real-estate video frames exhibiting forward-facing motions and room-scale translations.
\end{itemize}

\subsection{Evaluation Metrics}
\label{sec:supp_metrics}

We employ \textbf{Gemini-3-Pro} as an automated judge and assess each generated view along three complementary dimensions, each scored on a 1--5 Likert scale:

\begin{enumerate}[leftmargin=*,itemsep=2pt]
  \item \textbf{Instruction Compliance (IC)}: measures whether the generated image achieves the same camera viewpoint as the ground-truth target.  The evaluator receives the ground-truth target view and the generated view, and compares camera position, orientation, viewing angle, and the set of visible scene regions, while explicitly ignoring rendering style or image-quality differences.

  \item \textbf{Consistency with Original (CO)}: assesses whether objects and background elements present in the source image are faithfully preserved in the generated novel view.  The evaluator receives all three images (source, ground-truth target, and generated view), and checks each identifiable element for identity, appearance, and spatial arrangement consistency.

  \item \textbf{Rationality of New Content (RN)}: evaluates the plausibility of newly visible regions that the model must hallucinate (\eg, previously occluded surfaces or out-of-frame areas).  The evaluator judges whether the synthesized content exhibits coherent geometry, consistent materials and lighting, and seamless integration with the existing scene, using the ground-truth target as reference.
\end{enumerate}

\subsection{Training Convergence Analysis}
\label{sec:supp_convergence}

Table~\ref{tab:supp_convergence} reports evaluation scores at three training checkpoints, all fine-tuned from the same pre-trained T2I weights.  Key observations:

\begin{itemize}[leftmargin=*,itemsep=2pt]
  \item The 16k-step checkpoint achieves the highest overall score (2.773), with notably stronger Consistency with Original and Rationality of New Content than either the earlier or later checkpoints, suggesting that two epochs strike the best balance between learning geometric control and retaining generation quality.
  \item Extending training to 24k steps (${\sim}$3 epochs) leads to a clear drop in all three metrics compared with the 8k checkpoint, suggesting overfitting to the training distribution and degradation of generation diversity.
  \item Instruction Compliance remains relatively stable across all checkpoints (${\sim}$2.94--2.96), indicating that the model learns coarse viewpoint control early in training; the main gains from continued training manifest in finer-grained scene consistency and content plausibility.
\end{itemize}

\noindent
Based on these results, we select the 16k-step checkpoint as the final \joyomni{} model used throughout all experiments in the main paper.

\begin{table}[t]
  \centering
  \caption{\textbf{Training convergence analysis for \joyomni{}.}  We evaluate checkpoints at different training steps on our camera pose control benchmark (Section~\ref{sec:supp_benchmark}).  All models are fine-tuned from the pre-trained T2I weights.  \textbf{IC}: Instruction Compliance; \textbf{CO}: Consistency with Original; \textbf{RN}: Rationality of New Content.  All metrics are on a 1--5 scale ($\uparrow$ higher is better).  Best results are \textbf{bolded}.}
  \label{tab:supp_convergence}
  \vspace{4pt}
  \resizebox{0.5\columnwidth}{!}{
  \begin{tabular}{lcccc}
    \toprule
    \textbf{Checkpoint} & \textbf{IC\,$\uparrow$} & \textbf{CO\,$\uparrow$} & \textbf{RN\,$\uparrow$} & \textbf{Avg\,$\uparrow$} \\
    \midrule
    8k steps\;(${\sim}$1 epoch)   & \textbf{2.964} & 2.637 & 2.458 & 2.687 \\
    16k steps\;(${\sim}$2 epochs) & 2.941          & \textbf{2.696} & \textbf{2.683} & \textbf{2.773} \\
    24k steps\;(${\sim}$3 epochs) & 2.954          & 2.611 & 2.313 & 2.626 \\
    \bottomrule
  \end{tabular}
  }
\end{table}

\FloatBarrier 
\section{Capability Threshold Analysis}
\label{sec:supp_capability_threshold}

\begin{table}[htbp]
  \centering
  \caption{\textbf{Performance of \qwenSmall{} under our image-based paradigm.} Unlike the four stronger backbones evaluated in the main paper, \qwenSmall{} shows an overall accuracy \emph{decrease} after augmentation, suggesting the existence of a capability threshold. Values are accuracies (\%).}
  \label{tab:supp_capability_threshold}
  \resizebox{0.7\columnwidth}{!}{
  \begin{tabular}{lccccc}
  \toprule
  \textbf{Variant} & \textbf{Orientation} & \textbf{Location} & \textbf{Size} & \textbf{Multi-Object} & \textbf{Avg} \\
  \midrule
  Baseline & 43.1 & 64.8 & 80.0 & 56.9 & \textbf{56.5} \\
  \gemodels & 46.7 & 59.1 & 53.3 & 50.3 & 52.2 \\
  \bottomrule
  \end{tabular}
  }
  \vspace{-10pt}
  \end{table}

In the main paper (Section~\ref{sec:model}), we demonstrate that our image-based paradigm consistently improves all four evaluated understanding backbones, with the gain magnitude inversely correlated with baseline strength. A natural follow-up question is whether this trend extrapolates to even weaker models. To investigate, we additionally evaluate \qwenSmall{} under the identical generation and prompting protocol. Table~\ref{tab:supp_capability_threshold} reports the per-category and overall results.

In contrast to the consistent gains observed for the four stronger models, \qwenSmall{} exhibits an overall accuracy \emph{drop} of 4.3\,pp after augmentation (56.5\%\,$\to$\,52.2\%). While Orientation improves slightly (+3.6\,pp), Location, Size, and Multi-Object all degrade, with Size suffering the largest decline ($-$26.7\,pp).

To diagnose this anomalous behaviour, we analyse the failure-type distribution and compare it with that of a stronger model (\gpt{}). For \qwenSmall{}, the errors consist of 40.06\% \textit{Wrong Instruction} (133 cases), 42.47\% \textit{Bad Generation} (141 cases), and 17.47\% \textit{VL Failure} (58 cases), whereas the corresponding ratios for \gpt{} are 24.88\% (54), 60.83\% (132), and 14.29\% (31), respectively. Notably, \qwenSmall{} exhibits a substantially higher fraction of \textit{Wrong Instruction} errors than \gpt{}, indicating that a large portion of its failures arise from misinterpreting task requirements or failing to follow the intended instruction rather than from missing visual evidence.

This finding suggests the existence of a \emph{capability threshold}: the model must possess a basic level of task understanding (such as correctly parsing and executing the 3D-oriented instruction) before it can effectively benefit from the additional visual cues provided by novel-view synthesis. When instruction-following itself becomes the dominant failure mode, image-based augmentation provides limited headroom and may even amplify misaligned reasoning, explaining the negative overall effect observed for \qwenSmall{}. This analysis motivates the choice of the four backbones reported in the main paper and provides a practical guideline for practitioners: our paradigm is broadly beneficial across a wide capacity range, but the understanding backbone should meet a minimum instruction-following competence for the augmentation to take effect.

\section{Additional Qualitative Results}
\label{sec:additional_qualitative}

\begin{figure}[t]
\vspace{-15pt}
\centering
\includegraphics[width=\textwidth]{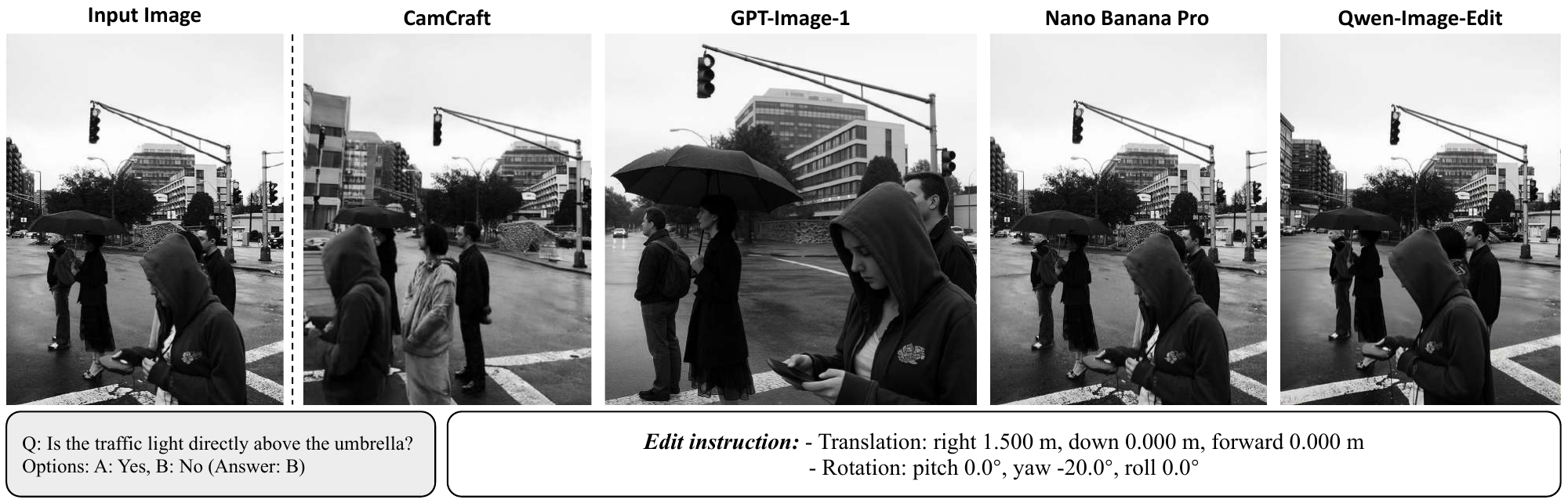}
\caption{\textbf{Qualitative comparison of novel-view synthesis across different generation models.} Given an input image and a camera transformation (right 1.5\,m, yaw $-20^{\circ}$), we compare the outputs of \joyomni{} (ours), GPT-Image-1, Nano Banana Pro, and Qwen-Image-Edit. While the competing models largely fail to faithfully follow the specified camera movement, producing outputs with incorrect viewpoint shifts, \joyomni{} accurately realizes the target transformation and preserves the original 3D spatial relationships.}
\label{fig:qualitative_comparison}
\vspace{-15pt}
\end{figure}

\begin{figure}[t]
\centering
\vspace{-15pt}
\includegraphics[width=\textwidth]{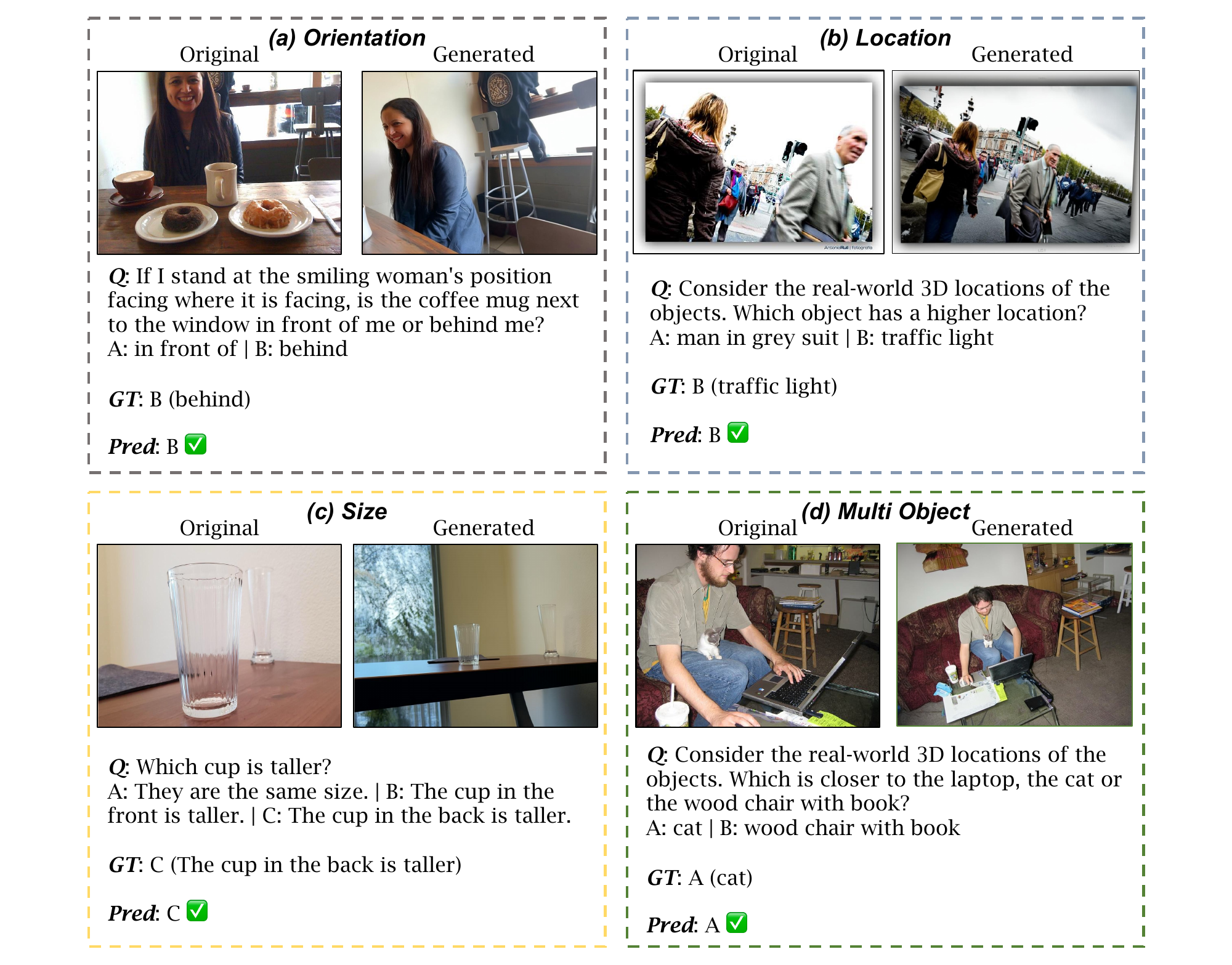}
\caption{\textbf{Qualitative examples of \joyomni{}-generated 
novel-view images across four evaluation categories.} 
For each category, namely (a)~Location, (b)~Orientation, (c)~Size,
and (d)~Multi-Object, we show the original image (left) and
the novel-view image synthesized by \joyomni{} (right) given 
non-trivial camera transformations (e.g., yaw up to  
$45^{\circ}$, translation up to $1.8$\,m). 
Despite the significant viewpoint changes, \joyomni{} 
consistently generates geometrically coherent images that 
preserve the underlying 3D spatial relationships. 
We verify this by querying Gemini-3-Flash on the 
synthesized views: the model correctly answers the 
corresponding spatial reasoning question in all four cases, 
confirming that \joyomni{} faithfully maintains the 3D cues 
required by each task type.}
\label{fig:qualitative_robustness}
\vspace{-15pt}
\end{figure}